\documentclass[10pt,twocolumn,letterpaper]{article}

\usepackage{arxiv}              
\usepackage{color}
\usepackage{xcolor} 
\usepackage{multirow}
\usepackage{pifont}
\usepackage{booktabs}
\usepackage{amsmath}
\usepackage{amssymb}
\usepackage{amsfonts}
\usepackage{subcaption}
\usepackage{makecell}
\usepackage{enumitem}
\usepackage{algorithmic}
\usepackage{amssymb}
\usepackage{pifont}
\makeatother

\usepackage[ruled,vlined]{algorithm2e}

\DeclareMathOperator*{\argmin}{arg\,min}
\definecolor{green}{HTML}{00800C}
\definecolor{blue}{HTML}{0064A6}
\definecolor{yellow}{HTML}{9EA310}

\definecolor{cvprblue}{rgb}{0.21,0.49,0.74}
\usepackage[pagebackref,breaklinks,colorlinks,allcolors=cvprblue]{hyperref}

\title{Q-DiT: Accurate Post-Training Quantization for Diffusion Transformers}

\author{
 Lei Chen$^{1}$ \quad Yuan Meng$^{1}$ \quad Chen Tang$^{1,2}$ \quad Xinzhu Ma$^{2}$ \quad Jingyan Jiang$^{3}$  \quad Xin Wang$^{1}$  \\  Zhi Wang$^{1}$ \quad Wenwu Zhu$^{1}$ \\
\normalsize $^{1}$Tsinghua University \quad $^{2}$MMLab, CUHK \quad $^{3}$Shenzhen Technology University
}

\begin{document}
\maketitle
\begin{abstract}
Recent advancements in diffusion models, particularly the architectural transformation from UNet-based models to Diffusion Transformers (DiTs), significantly improve the quality and scalability of image and video generation. However, despite their impressive capabilities, the substantial computational costs of these large-scale models pose significant challenges for real-world deployment. Post-Training Quantization (PTQ) emerges as a promising solution, enabling model compression and accelerated inference for pretrained models, without the costly retraining. However, research on DiT quantization remains sparse, and existing PTQ frameworks, primarily designed for traditional diffusion models, tend to suffer from biased quantization, leading to notable performance degradation. In this work, we identify that DiTs typically exhibit significant spatial variance in both weights and activations, along with temporal variance in activations. To address these issues, we propose Q-DiT, a novel approach that seamlessly integrates two key techniques: automatic quantization granularity allocation to handle the significant variance of weights and activations across input channels, and sample-wise dynamic activation quantization to adaptively capture activation changes across both timesteps and samples. Extensive experiments conducted on ImageNet and VBench demonstrate the effectiveness of the proposed Q-DiT. Specifically, when quantizing DiT-XL/2 to W6A8 on ImageNet ($256 \times 256$), Q-DiT achieves a remarkable reduction in FID by 1.09 compared to the baseline. Under the more challenging W4A8 setting, it maintains high fidelity in image and video generation, establishing a new benchmark for efficient, high-quality quantization in DiTs. 
Code is available at \href{https://github.com/Juanerx/Q-DiT}{https://github.com/Juanerx/Q-DiT}. 
\end{abstract}    
\begin{figure*}[t]
    \centering
    \includegraphics[width=0.95\textwidth]{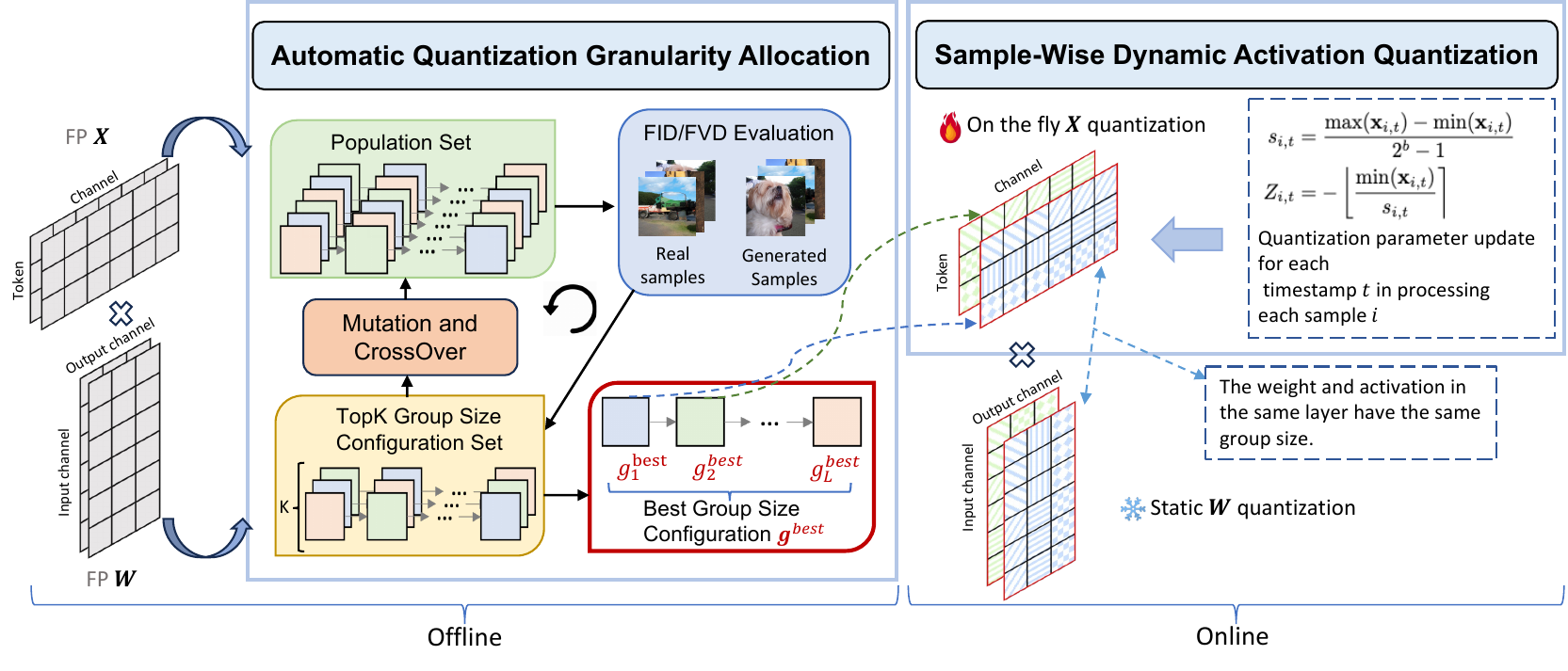}
  \caption{Overview of the proposed Q-DiT.
   The weights and activations within each layer are quantized with the same group size. Group size configurations allocated for each layer are based on the evolutionary search results, which are guided by the FID/FVD score between the real samples and samples generated by the quantized model. The activations are dynamically quantized during runtime. 
  }
  \label{fig: overview}
\end{figure*}

\section{Introduction}
Diffusion models \citep{nichol2021improved,ho2020denoising, songddim,diffusion_survey} have emerged as a powerful base model for various tasks, ranging from computer vision, natural language processing, multi-modal modeling, {\it etc.} The architectural design of diffusion models has evolved significantly. Traditionally, these models employed UNet~\citep{ronneberger2015unet} architecture due to their efficiency in managing hierarchical feature representations. However, recent advances have shifted the focus towards diffusion transformers (DiTs)~\citep{peebles2023scalable}, and notable examples, including Stable Diffusion 3~\citep{esser2024scaling} and Sora~\citep{sora}, have demonstrated its superior performance and scalability for complex generative tasks. 

Despite their success, a significant limitation of DiTs lies in their inherently high latency in the inference. The iterative denoising process, although effective, requires numerous sampling steps, making real-time or large-scale applications computationally intensive. Model quantization offers a particularly promising avenue to reduce inference latency, and Post Training Quantization (PTQ) is particularly appealing for large models as it eliminates the need for retraining.
However, the application of quantization techniques to transformer-based diffusion models remains limited. Existing quantization methods for diffusion models~\citep{shang2023ptqdm, he2024ptqd, he2023efficientdm} primarily focus on UNet architecture and heavily rely on reconstruction-based methods, challenging to scale to large models~\citep{nagel2020up, li2021brecq}.

In this work, we aim to propose a customized quantization method for DiTs. To achieve this, we first explore the distinct characteristics of DiT models and identify two key issues in DiT quantization: \emph{significant variance of weights and activations across input channels} and \emph{varying activations across different timesteps}.
Therefore, we propose Q-DiT, consisting of a fine-grained group quantization strategy and a dynamic activation quantization strategy. These two designs address the aforementioned challenges individually and collaboratively contribute to the proposed quantization framework (see Fig. \ref{fig: overview}).

In particular, for the first challenge, a promising solution is group quantization \citep{zhao2024atom,lin2023awq} which can manage high-magnitude values at the group level. However, we observe the non-monotonicity in group sizes, {\it e.g.,} reducing group size (increasing group number) does not always lead to better performance. Consequently, we employ an evolutionary search algorithm to configure group sizes for quantization across different model layers. 
This method utilizes the Fréchet Inception Distance (FID) and Fréchet Video Distance (FVD) as metrics to directly correlate the quantization effects with the visual quality of generated samples, enabling a more targeted and effective quantization strategy. 
The evolutionary approach not only identifies the optimal group sizes but also ensures that the quantization process adheres to predefined computational constraints, effectively balancing performance with efficiency.

Furthermore, varying activations across the timesteps, the second challenge, indicates that quantization parameters calibrated at specific timesteps may not generalize well in all timesteps. To address this variability, Q-DiT adopts a sample-wise dynamic activation quantization mechanism, which adapts with sample granularity to the changing distribution of activations throughout the diffusion process. This approach significantly reduces quantization error by adjusting quantization parameters on-the-fly, ensuring high-quality image/video generation with minimal overhead.

In summary, our main contributions are as follows:
\begin{itemize}[leftmargin=*]
\setlength{\itemsep}{1pt}
\item We introduce Q-DiT, an accurate PTQ method designed for DiTs. This method employs fine-grained group quantization to effectively manage input channel variance in both weights and activations, and it adopts sample-wise dynamic activation quantization to adapt to activation variations across different timesteps and samples.
\item We identify that the default group size configuration is sub-optimal and propose an evolutionary search strategy to optimize group size allocation, which enhances the efficiency and efficacy of the quantization process.
\item Extensive experiments on ImageNet and VBench demonstrate that Q-DiT achieves lossless compression under a W6A8 configuration and minimal degradation under W4A8 for image and video generation, highlighting its superior performance. 
\end{itemize}
\section{Related Work}
\label{relatedwork}
\noindent\textbf{Model quantization.} 
Model quantization is a widely used technique to reduce the model size and accelerate its inference speed by converting the model's weights and activations from high-precision floating-point numbers to lower-precision numbers. 
There are two primary approaches to quantization: Quantization-Aware Training (QAT)~\citep{choi2018pact, esser2019learned, bhalgat2020lsq+} and Post-Training Quantization (PTQ)~\citep{nagel2020up, li2021brecq}. QAT integrates the quantization process directly into the fine-tuning phase, leveraging STE~\citep{bengio2013estimating} to simultaneously optimize quantizer parameters and model parameters during fine-tuning. This approach restores the model's performance degradation caused by quantization. However, QAT is resource-intensive because it necessitates fine-tuning the model on the original training dataset. In contrast, PTQ is far more efficient and practical, as it does not require model retraining. PTQ operates by utilizing a small calibration dataset to adjust the quantization parameters for weights and activations, facilitating significant model compression with minimal effort. Although PTQ is highly efficient, it can result in significant performance degradation when applied to low-bit quantization. Reconstruction-based method~\citep{nagel2020up, li2021brecq} tries to minimize performance degradation by reducing the reconstruction error of each layer or each block. Although the reconstruction-based method performs well in CNN, they are not easy to scale up to a large model.

\noindent\textbf{Quantization of transformers.} 
Quantization of transformers has been extensively researched in the contexts of both Vision Transformers (ViTs) and Large Language Models (LLMs). Specifically, PTQ4ViT~\citep{yuan2022ptq4vit} proposed the twin uniform quantizer to handle the special distributions of post-softmax and post-GELU activations. RepQ-ViT~\citep{li2023repq} used scale reparameterization to reduce the quantization error of activations. For LLM, weight-only quantization quantizes the weight to reduce the heavy memory movements to achieve better inference efficiency. GPTQ~\citep{frantar2022gptq} reduced the bit-width to 4 bits per weight based on approximate second-order information with weight-only quantization. 
AWQ~\citep{lin2023awq} proposed activation-aware weight quantization to reduce the quantization error of salient weight. 
On the other hand, weight-activation quantization further enhances inference efficiency by quantizing both weight and activation but has to face activation outliers. 
LLM.\texttt{int8()}~\citep{dettmers2022gpt3} reduces the effect of outliers by keeping them in \texttt{FP16} with mixed-precision computations. Outlier Suppression~\citep{wei2022outlier} reduces the quantization error of activations by using the non-scaling LayerNorm. 
However, these quantization techniques may not be directly applied to DiTs, due to their diffusion model characteristics.

\noindent\textbf{Quantization of diffusion models.} 
Diffusion models tend to have a slow inference speed due to the large number of sampling steps required. Consequently, some recent studies have focused on accelerating these models through quantization techniques. PTQ4DM~\citep{shang2023ptqdm} and Q-diffusion~\citep{li2023q} discover activation variance across different denoising steps and adopt reconstruction-based methods for quantization. PTQD~\citep{he2024ptqd} finds the correlation between the quantization noise and model output and proposes variance schedule calibration to rectify the uncorrelated part. 
TDQ~\citep{so2024temporal} utilizes an MLP layer to estimate the activation quantization parameters for each step. 
TMPQ-DM~\citep{sun2024tmpq} further reduces the sequence length of timestep along with the quantization to reduce the overall costs. PTQ4DiT~\citep{wu2024ptq4dit} introduces a PTQ method tailored for Diffusion Transformers, addressing challenges like extreme channel magnitudes and temporal activation variability using Channel-wise Salience Balancing (CSB) and Spearman’s $\rho$-guided Salience Calibration (SSC), while achieving W4A8 quantization.
Comparison table of PTQ methods for diffusion models can be found in the supplementary materials.
These methods are unable to handle simultaneously the characteristics of the transformer architecture and the dynamics of activation during denoising process, leading to significant performance drops.

\section{Observations of DiT Quantization}
\label{challenge}
We find directly applying recent UNet-based quantization methods to quantize DiTs will lead to significant performance degradation. To understand the underlying reasons, we explore the distinct characteristics of DiT models, particularly how they differ from UNet-based architectures in terms of weight and activation distributions.

\begin{figure}[t]
  \centering
  \begin{subfigure}{\columnwidth}
    \includegraphics[width=\columnwidth]{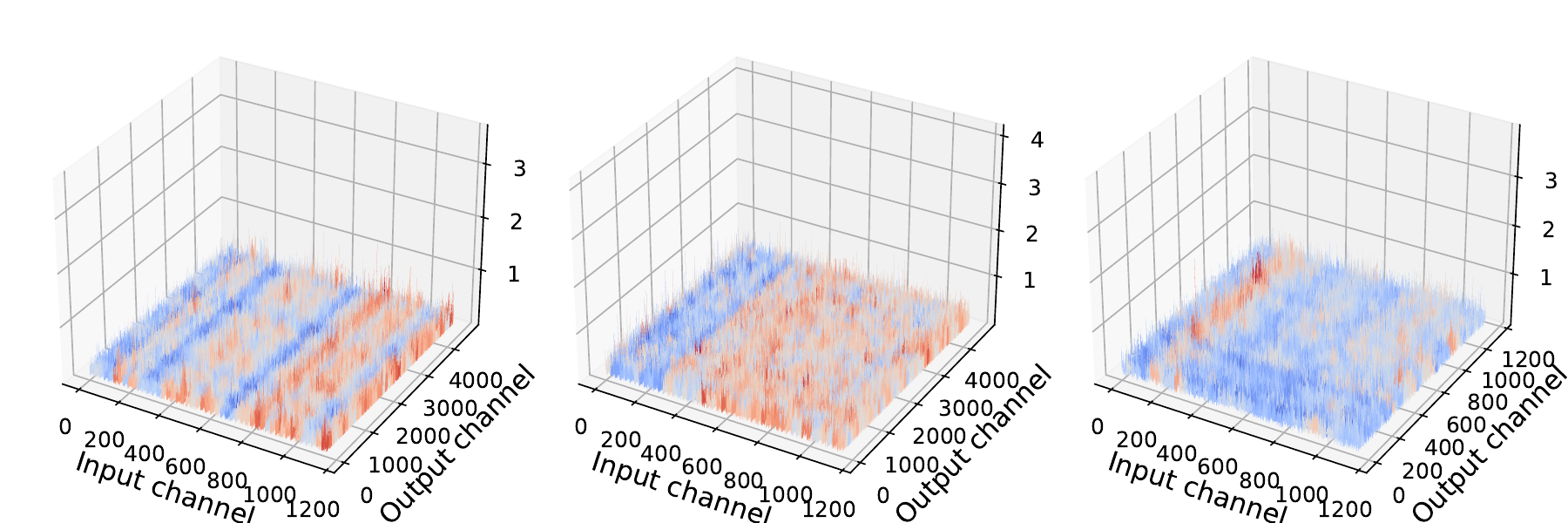}
    \caption{\textbf{Weight}  distribution}
    \label{fig:weight_3d}
  \end{subfigure}
  \hfill
  \begin{subfigure}{\columnwidth}
    \includegraphics[width=\columnwidth]{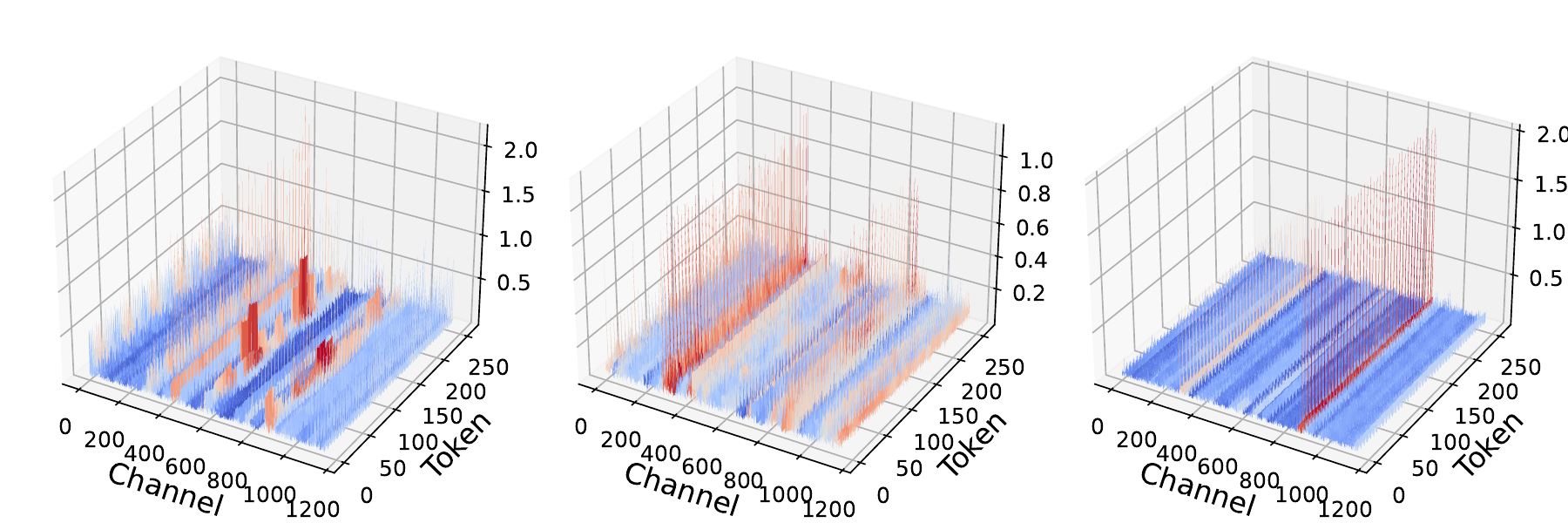}
    \caption{\textbf{Activation} distribution}
    \label{fig:activation_3d}
  \end{subfigure}
  \caption{Distributions of weights and activations in different layers of DiT-XL/2. The red peaks indicate higher values, while the blue areas represent lower values.}
  \label{fig:3d_plot}
\end{figure}

\noindent 
\textbf{Observation 1}: \emph{DiTs exhibit significant variance of weights and activations across input channels.}
As shown in Fig.~\ref{fig:weight_3d}, the variance of weights and activations across input channels is much more significant than the output channels.
This variance substantially affects quantization since common methods typically apply channel-wise quantization along the output channel for diffusion models~\citep{shang2023ptqdm, he2024ptqd}.
Besides, we also find outliers persist in specific channels of the activation, as shown in Fig.~\ref{fig:activation_3d}. This suggests that, if we continue to use tensor-wise quantization, these outliers will significantly impact the quantization parameters, resulting in substantial quantization errors for non-outliers. 

\begin{figure}[htbp]
    \centering
    \includegraphics[width=1\columnwidth]{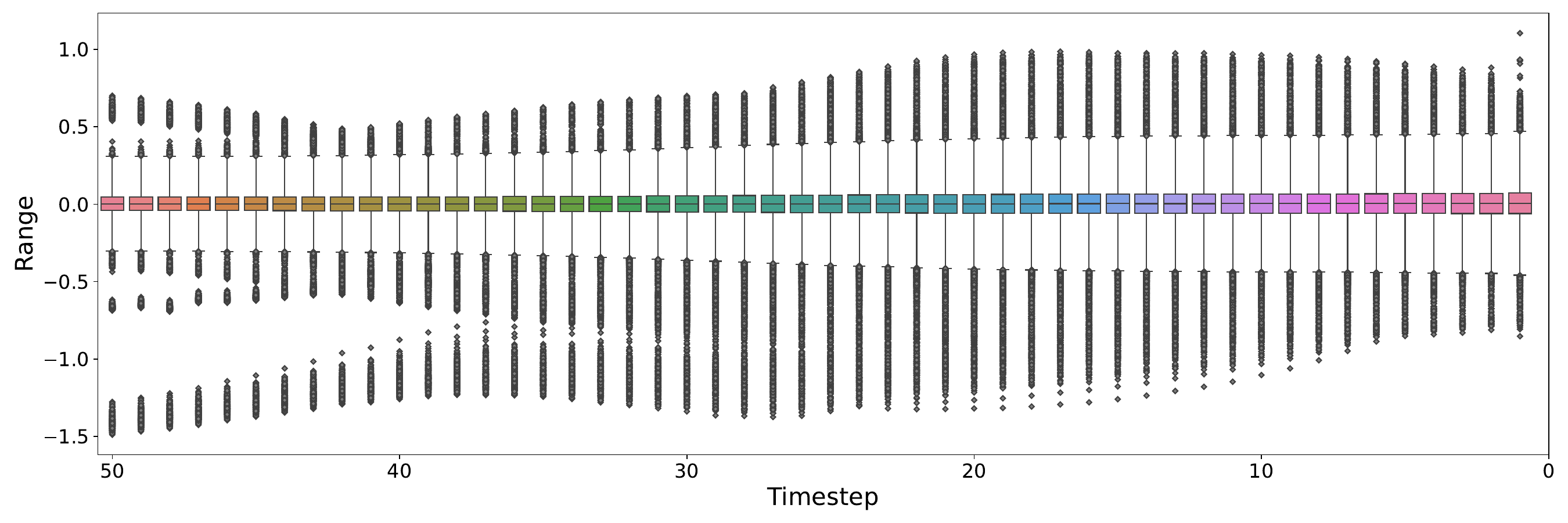}
  \caption{
  Box plot showing the distribution of activation values across various timesteps (from 50 to 0) for the DiT-XL/2 model when
generating one image from ImageNet at $256 \times 256$ resolution..
  }
  \label{fig: actdis}
\end{figure}

\begin{figure}[htbp]
\centering
\includegraphics[width=1\columnwidth]{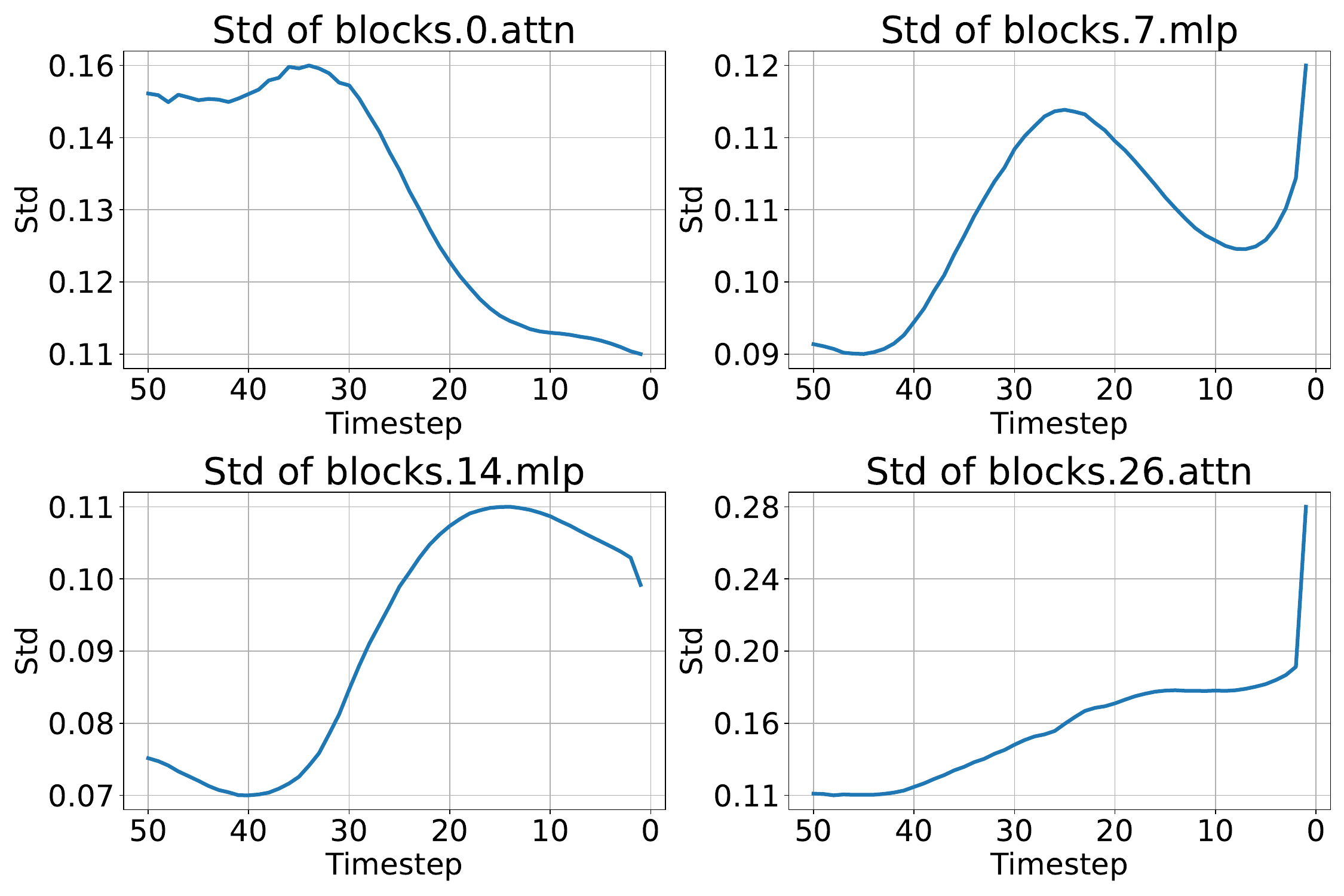}
\caption{Standard deviation of activations in MLP and attention layers across different blocks over 50 timesteps for DiT-XL/2 when
generating one image from ImageNet at $256 \times 256$ resolution.}
\label{fig:actstd}
\end{figure}

\noindent \textbf{Observation 2}: \emph{Significant distribution shift of activations across timesteps.}  We observe that the distribution of activations in DiT models undergoes significant changes at different timesteps during the denoising process, as demonstrated in Fig.~\ref{fig: actdis} and Fig.~\ref{fig:actstd}. 
Further, we discovered that this temporal shift also exhibits significant variability across different samples.
The relevant experimental results are presented in the supplementary materials.

\section{Preliminary}
We use uniform quantization to quantize both weights and activations in this work, as it is more hardware-friendly~\citep{jacob2018quantization, han2015deep}. Particularly, uniform quantization divides the range of floating-point values into equally spaced intervals, and each interval is mapped to a discrete value. The uniform quantization function $Q$ that quantize input floating-point tensor $\mathbf{x}$ into $b$ bit integer tensor $\hat{\mathbf{x}}$ can be expressed as: 
\begin{equation} \label{eq:quant}
\hat{\mathbf{x}}=Q(\mathbf{x}; b)=s \cdot (\mathrm{clip}(\lfloor \frac{\mathbf{x}}{s} \rceil + Z, 0, 2^{b}-1)-Z),
\end{equation}
where $s = \frac{\max(\mathbf{x}) - \min(\mathbf{x})}{2^b - 1} ~\text{and}~ Z = - \left\lfloor \frac{\min(\mathbf{x})}{s} \right\rceil$.
Here, $\lfloor \cdot \rceil$ is the $\mathrm{round}$ operation, $s$ is the scaling factor and $Z$ denotes the zero point. 

\section{Method: Q-DiT}
\label{qdit}

As shown in Fig. \ref{fig: overview}, the proposed Q-DiT involves two novel designs, including automatic quantization granularity allocation and dynamic activation quantization. Here we introduce them in the following parts.

\subsection{Automatic Quantization Granularity Allocation}
{\bf Base solution.} A straightforward solution to deal with the input channel-wise variance, as highlighted in {\bf observation 1}, is to apply input channel-wise quantization, using different quantization parameters for each channel.
However, this approach compromises computational efficiency during hardware deployment, as it prevents the full utilization of low-precision computation due to the need for repeated intermediate rescaling ~\citep{bondarenko2021understanding, xiao2023smoothquant}. 

\noindent \textbf{Fine-grained group quantization.} As discussed in recent LLM quantization work~\citep{zhao2024atom,lin2023awq}, a compromise approach between input channel-wise quantization and tensor quantization is the group quantization. As shown in Fig.~\ref{fig: overview}, the weight and activation matrices are divided into groups, and then we perform quantization for each group separately. Specifically, consider a matrix multiplication $\mathbf{Y}=\mathbf{X}\mathbf{W}$ in a linear layer, where $\mathbf{X} \in \mathbb{R}^{n \times d_{\text{in}}}$ and $\mathbf{W} \in \mathbb{R}^{d_{\text{in}} \times d_{\text{out}}}$. The quantized value for each output element can be obtained by:
\begin{align} \label{eq:group}
\hat{\mathbf{Y}}_{i,j} &= \sum_{k=0}^{d_{\text{in}}} \hat{\mathbf{X}}_{i,k}\hat{\mathbf{W}}_{k,j} \\
&= \sum_{u=0}^{d_{\text{in}}/g_{ll}-1}\sum_{v=0}^{g_{ll}} Q_{u}^{\mathbf{X}}({\mathbf{X}}_{i,ug_{ll}+v})Q_{u}^{\mathbf{W}}({\mathbf{W}}_{ug_{ll}+v,j}),
\end{align}
where $g_{ll}$ denotes the group size.

\noindent \textbf{Non-monotonicity in quantization group selection. } 
Ideally, we can improve model performance by reducing the group size ({\it i.e.,} increase the group number) because finer-grained quantization reduces quantization error. However, as shown in Tab.~\ref{_tab:group_observation}, we have observed that smaller group sizes do not always yield better results, where such existence of non-monotonicity in the quantization group further demonstrates that DiT quantization is rather different compared with LLM and ViT quantizations. For instance, when the group size reduces from 128 to 96, the FID increases by about 11.8\%, from 17.87 to 19.97, indicating a degradation in the quality of generated images. 
This suggests that there is an optimal group size configuration that can achieve better quantization effects with the same average group size or achieve the same quantization results with a larger average group size. 
Additionally, the sensitivity of each layer in the model varies. By assigning different group sizes to different layers, we can achieve high efficiency and quality in both model performance and image generation. 

\begin{table}[t]
\caption{Quantization results with varying group sizes on ImageNet 256$\times$256 and 512$\times$512.}
\centering
\small
\resizebox{0.8\linewidth}{!}{
\begin{tabular}{ccc|ccc}
\toprule
\multicolumn{3}{c|}{256$\times$256} & \multicolumn{3}{c}{512$\times$512}                   \\ \midrule
Group  & FID $\downarrow$    & sFID $\downarrow$  & Group & FID $\downarrow$   & \multicolumn{1}{r}{sFID $\downarrow$} \\ \midrule
128         & 17.87  & 20.45 & 96         & 20.76 & 21.97                    \\
96          & 19.97  & 21.42 & 64         & 20.90 & 22.58         \\ \bottomrule
\end{tabular}}
\label{_tab:group_observation}
\end{table} 

\noindent \textbf{Automatic group allocation.} 
The primary challenge in allocating group sizes lies in identifying the correlation between the group size of each layer and the final generation performance of the diffusion model. Previous works on mixed precision quantization focus on identifying sensitivity indicators~\citep{li2021brecq, tang2022mixed} for each layer, such as the MSE between the quantized layer and the full precision layer, and then transforming this into an integer linear programming (ILP) problem for optimization. However, we find that a smaller MSE does not necessarily correspond to a reduced performance degradation for DiT quantization, indicating that the previous methods may be ineffective.

To this end, we directly use the FID as our metric for image generation, defined as follows:
\begin{equation}
\label{Eq2}
    L(\mathbf{g}) = \text{FID}(R, G_{\mathbf{g}}).
\end{equation}
Similarly, for video generation models, we use the FVD as our metric:
\begin{equation}
\label{Eq3}
    L(\mathbf{g}) = \text{FVD}(R, G_{\mathbf{g}}),
\end{equation}
where $\mathbf{g} = \{g_1, g_2, \dots,
g_N\}$
 is the layer-wise group size configuration and $N$ is the number of quantized layers. 
 \(R\) and \(G_{\mathbf{g}}\) denote the real samples and the samples generated by the quantized model, respectively.
We then employ an evolutionary algorithm to optimize the following objective function:
\begin{equation}
\label{Eq4}
    {\bf{g}}^{\ast} = \argmin_{\mathbf{g}} L(\mathbf{g}) , \ \ \text{s.t. } B({\bf{g}}) \le N_{bitops},
\end{equation}
where \(B(\cdot)\) is the measurement of bit-operations (BitOps), and \(N_{bitops}\) is the predefined threshold. 

This approach allows us to better capture the nuanced impacts of group size on quantization performance, leading to improved outcomes in both efficiency and image quality. The algorithm is located in Alg.~\ref{alg:alg1}.

\begin{algorithm}[t!]
    \caption{Automatic quantization granularity allocation of Q-DiT}
    \label{alg:alg1}
    \KwIn{Group size search space $\mathcal{S}_g$; number of layers $L$; population size $N_p$; iterations $N_{iter}$; mutation probability $p$; Constraint $N_{bitops}$}
    Initialize population $\mathcal{P}=\{\mathbf{g}^j\}_{j=1}^{N_p}$, where each element in configuration $\mathbf{g}^j \in \mathbb{R}^L$ is  randomly selected from $\mathcal{S}_g$;\\
    Initialize TopK candidate set $\mathcal{S}_{\text{TopK}}= \emptyset$\;
    \For{$t=1,2,\dots, N_{iter}$}
          {
            \For{$i=1,2,\dots, N_{p}$}
            {Calculate FID (or FVD) for each configuration $\mathbf{g}^j$ based on Eq.~\ref{Eq2} (or Eq.~\ref{Eq3})\;}
            Update $\mathcal{S}_{\text{TopK}}$ with $K$ configurations, according to ranked FID (or FVD) scores\;
            Clear population $\mathcal{P} = \emptyset$\;
            \Repeat{$|\mathcal{P}|$ = $N_{p}/2$          
            }{
                $\mathbf{g}_{cross} = $ CrossOver($\mathcal{S}_{\text{TopK}}$) with probability $1-p$\;
                Append $\mathbf{g}_{cross}$ to $\mathcal{P}$ if $B(\mathbf{g}_{cross})<N_{bitops}$\;
            }
            \Repeat{ $|\mathcal{P}|$ = $N_{p}$          
            }{
                $\mathbf{g}_{mutate} = $ Mutate($\mathcal{S}_{\text{TopK}}$) with probability $p$\;
                Append $\mathbf{g}_{mutate}$ to $\mathcal{P}$ if $B(\mathbf{g}_{mutate})<N_{bitops}$\;
            }
          }
    Get the best group size configuration $\mathbf{g}^{\text{best}}$, and use it to quantize the model\;
    \textbf{return} quantized model
\end{algorithm}

\subsection{Sample-wise Dynamic Activation Quantization}
In UNet-based diffusion quantization, previous studies either allocate a set of quantization parameters for all activations at each timestep~\citep{he2023efficientdm} or design a multi-layer perceptron (MLP)~\citep{so2024temporal} to predict quantization parameters based on the timestep. However, due to {\bf observation 2}, these methods are not compatible with our fine-grained quantization because assigning quantization parameters to each group at every timestep results in considerable memory overheads. Specifically, for a sampler with 50 timesteps, the memory overhead could reach up to 39\% of the full-precision model's size.

Inspired by the recent work in LLM optimization~\citep{wu2023zeroquant}, we design an on-the-fly dynamic quantization approach for activations. 
Specifically, during inference, the quantization parameters for each group of the activations are calculated based on their {\tt min-max} values. For a given sample $i$ at timestep $t$, the quantization parameters for the activation $\mathbf{x}_{i,t}$ can be expressed as:
\begin{align} \label{eq:Eq4}
s_{i,t} &= \frac{\max(\mathbf{x}_{i,t}) - \min(\mathbf{x}_{i,t})}{2^b - 1}, \\
Z_{i,t} &= - \left\lfloor \frac{\min(\mathbf{x}_{i,t})}{s_{i,t}} \right\rceil.
\end{align}
Furthermore, we integrate the dynamic quantization with {\tt min-max} computation into the prior operator, which can benefit from the operator fusion and the overhead becomes negligible compared to the costly matrix multiplications in transformer blocks. 
\begin{table*}[t!]
\caption{Results on image generation. We show the quantization results of DiT-XL/2 on ImageNet 256$\times$256 and 512$\times$512. 'W/A' indicates the bit-width of weight and activation, respectively.}
\vspace{-3pt}
\centering
\label{tab: 256_performance}
{%
\small
\resizebox{0.72\linewidth}{!}{
\begin{tabular}{cccccccc}
\toprule
Model             & \begin{tabular}[c]{@{}c@{}}Bit-width (W/A)\end{tabular} & Method & Size (MB) & FID $\downarrow$ & sFID $\downarrow$ & IS $\uparrow$ & Precision $\uparrow$ \\ \midrule
\multirow{14}{*}{\parbox{2cm}{\centering DiT-XL/2\\ 256$\times$256 \\ (steps = 100)}} & \textcolor{gray}{16/16}                                   & \textcolor{gray}{FP}     & \textcolor{gray}{1349}    & \textcolor{gray}{12.40} & \textcolor{gray}{19.11} & \textcolor{gray}{116.68} & \textcolor{gray}{0.6605} \\ \cmidrule{2-8} 
                      & \multirow{6}{*}{6/8}     & PTQ4DM   &508 &17.86 &25.33  &92.24  &0.6077  \\
                      &                          & RepQ-ViT  &508  &27.74  &20.91  &63.41  &0.5600  \\
                      &                          & TFMQ-DM   &508 &22.33 &27.44  &72.74  &0.5869  \\
                      &                          & PTQ4DiT  &508  &15.21  &21.34  &105.03  &0.6440  \\
                      &                          & G4W+P4A      &520  &16.72  &24.61  &100.09  &0.6123  \\
                      &                          & \textbf{Ours}   &518     &\textbf{12.21}  &\textbf{18.48}  &\textbf{117.75}  &\textbf{0.6631}   \\ \cmidrule{2-8} 
                      & \multirow{6}{*}{4/8}     & PTQ4DM   &339   &213.66  &85.11  &3.26  &0.0839 \\
                      &                          & RepQ-ViT   &339   &224.14  &81.24  &3.25  &0.0373  \\
                      &                          & TFMQ-DM   &339   &143.47  &61.09  &5.61  &0.0497 \\
                      &                          & PTQ4DiT   &339   &28.90  &34.56  &65.73  &0.4931  \\
                      &                          & G4W+P4A      &351  &25.48  &25.57  &73.46  &0.5392  \\
                      &                          & \textbf{Ours}  &347 &\textbf{15.76}  &\textbf{19.84}   &\textbf{98.78}  &\textbf{0.6395}  \\ \midrule

\multirow{14}{*}{\parbox{2cm}{\centering DiT-XL/2 \\ 256$\times$256 \\ (steps = 100 \\ cfg = 1.5)}}  & \textcolor{gray}{16/16}                                   & \textcolor{gray}{FP}     & \textcolor{gray}{1349}     & \textcolor{gray}{5.31} & \textcolor{gray}{17.61} & \textcolor{gray}{245.85} & \textcolor{gray}{0.8077} \\ \cmidrule{2-8} 
                      & \multirow{6}{*}{6/8}     & PTQ4DM &508      &8.41  &25.56  &196.73 &0.7622  \\
                      &                          & RepQ-ViT  &508    &10.77  &18.53  &163.11  &0.7264  \\
                      &                          & TFMQ-DM &508      &8.87  &23.52  &194.08 &0.7737  \\
                      &                          & PTQ4DiT  &508    &5.34  &18.48  &209.90  &\textbf{0.8047}  \\
                      &                          & G4W+P4A      &520  &6.41  &19.52  &225.48  &0.7705  \\
                      &                          & \textbf{Ours}    &518  &\textbf{5.32}  &\textbf{17.40}  &\textbf{243.95}  &0.8044  \\ \cmidrule{2-8} 
                      & \multirow{6}{*}{4/8}     & PTQ4DM   &339  &215.68 &86.63  &3.24 &0.0741  \\
                      &                          & RepQ-ViT     &339   &226.60  &77.93  &3.61  &0.0337  \\
                      &                          & TFMQ-DM   &339  &141.90 &56.01  &6.24 &0.0439  \\
                      &                          & PTQ4DiT     &339   &7.75  &22.01  &190.38  &0.7292  \\
                      &                          & G4W+P4A      &351  &7.66  &20.76  &193.76  &0.7261  \\
                      &                          & \textbf{Ours}    &347    &\textbf{6.40}  &\textbf{18.60}  &\textbf{211.72} &\textbf{0.7609} \\ \midrule
\multirow{14}{*}{\parbox{2cm}{\centering DiT-XL/2 \\ 512$\times$512 \\ (steps = 50)}} & \textcolor{gray}{16/16}                                   & \textcolor{gray}{FP}     & \textcolor{gray}{1349}    & \textcolor{gray}{16.01} & \textcolor{gray}{20.50} & 
\textcolor{gray}{97.79} & \textcolor{gray}{0.7481} \\ \cmidrule{2-8} 
                      & \multirow{6}{*}{6/8}     & PTQ4DM &508      &21.22  &20.11  &80.07  &0.7131  \\
                      &                          & RepQ-ViT  &508    &19.67  &22.35  &75.78  &0.7082  \\
                      &                          & TFMQ-DM &508      &20.99  &22.01  &71.08  &0.6918  \\
                      &                          & PTQ4DiT  &508    &19.42  &21.94  &77.35  &0.7024  \\
                      &                          & G4W+P4A      &520  &19.55  &22.43  &85.56  &0.7158  \\
                      &                          & \textbf{Ours}    &517  &\textbf{16.21}  &\textbf{20.41}  &\textbf{96.78}  &\textbf{0.7478}  \\ \cmidrule{2-8} 
                      & \multirow{6}{*}{4/8}     & PTQ4DM   &339   &131.66  &75.79  &11.54  &0.1847 \\
                      &                          & RepQ-ViT   &339   &105.32  &65.63  &18.01  &0.2504  \\
                      &                          & TFMQ-DM   &339   &80.70  &59.34  &29.61  &0.2805 \\
                      &                          & PTQ4DiT   &339   &35.82  &28.92  &48.62  &0.5864  \\
                      &                          & G4W+P4A      &351  &26.58  &24.14  &70.24  &0.6655  \\
                      &                          & \textbf{Ours}   &348    &\textbf{21.59}  &\textbf{22.26}   &\textbf{81.80}  &\textbf{0.7076}  \\ \midrule

\multirow{14}{*}{\parbox{2cm}{\centering DiT-XL/2 \\ 512$\times$512 \\ (steps = 50 \\ cfg = 1.5)}}  & \textcolor{gray}{16/16}                                   & \textcolor{gray}{FP}     & \textcolor{gray}{1349}     & \textcolor{gray}{6.27} & 
\textcolor{gray}{18.45} & \textcolor{gray}{204.47} & \textcolor{gray}{0.8343} \\ \cmidrule{2-8} 
                      & \multirow{6}{*}{6/8}     & PTQ4DM &508      &9.84  &26.57  &164.91  &0.8215  \\
                      &                          & RepQ-ViT  &508    &8.30  &19.19  &158.80  &0.8153  \\
                      &                          & TFMQ-DM &508      &8.34  &17.94  &162.16  &0.8262  \\
                      &                          & PTQ4DiT  &508    &7.69  &18.86  &178.34  &0.8121  \\
                      &                          & G4W+P4A      &520  &7.28  &19.62  &185.92  &0.8143  \\
                      &                          & \textbf{Ours}    &517   &\textbf{6.24}  &\textbf{18.36}  &\textbf{202.48}  &\textbf{0.8341}  \\ \cmidrule{2-8} 
                      & \multirow{6}{*}{4/8}     & PTQ4DM   &339  &88.45 &50.80  &26.79 &0.3206  \\
                      &                          & RepQ-ViT     &339   &79.69  &49.76  &29.46  &0.3413  \\
                      &                          & TFMQ-DM   &339  &54.61 &44.27  &58.77 &0.4215  \\
                      &                          & PTQ4DiT     &339   &11.69  &22.86  &117.34  &0.7121  \\
                      &                          & G4W+P4A      &351  &9.98  &20.76  &156.07  &0.7840  \\
                      &                          & \textbf{Ours}  &347  &\textbf{7.82}  &\textbf{19.60}  &\textbf{174.18} &\textbf{0.8127} \\ \bottomrule
\end{tabular}}
}
\vspace{-0.3cm}
\end{table*}

\begin{table*}[h!]
\caption{Results on video generation. We show the quantization results of STDiT3 on VBench. Higher metrics indicate better performance.}
\label{tab:video_model_performance}
\vspace{-3pt}
\centering
\small
\resizebox{0.93\linewidth}{!}{
\begin{tabular}{@{}cccccccccc@{}}
\toprule
Method & \begin{tabular}[c]{@{}c@{}}Bit-width\\(W/A)\end{tabular} & \makecell{Subject\\Consistency} & \makecell{Overall\\Consistency}  & \makecell{Temporal\\Style} & \makecell{Appearance\\Style}  & Scene & \makecell{Spatial\\Relationship} & Color & \makecell{Human\\Action} \\ \midrule
\textcolor{gray}{FP} &\textcolor{gray}{16/16} & \textcolor{gray}{0.9522} & \textcolor{gray}{0.2667} & \textcolor{gray}{0.2507} & \textcolor{gray}{0.2352} & \textcolor{gray}{0.4094} & \textcolor{gray}{0.3441} & \textcolor{gray}{0.7864} & \textcolor{gray}{0.8680} \\ \midrule
G4W+P4A & 4/8 & 0.9444 & 0.2628 & 0.2489 & 0.2344 & \textbf{0.3924} & 0.3265 & 0.7657 & 0.8600 \\
\textbf{Ours} & 4/8 & \textbf{0.9498} & \textbf{0.2663} & \textbf{0.2511} & \textbf{0.2346} & 0.3871 & \textbf{0.3810} & \textbf{0.7947} & \textbf{0.8620} \\
\bottomrule
\toprule
Method & \begin{tabular}[c]{@{}c@{}}Bit-width\\(W/A)\end{tabular} & \makecell{Multiple\\Objects} & \makecell{Object\\Class} & \makecell{Imaging\\Quality} & \makecell{Aesthetic\\Quality} & \makecell{Dynamic\\Degree} & \makecell{Motion\\Smoothness} & \makecell{Temporal\\Flickering} & \makecell{Background\\Consistency} \\ \midrule
\textcolor{gray}{FP} &\textcolor{gray}{16/16} & \textcolor{gray}{0.4143} & \textcolor{gray}{0.8383} & \textcolor{gray}{0.5829} & \textcolor{gray}{0.5173} & \textcolor{gray}{0.6139} & \textcolor{gray}{0.9855} & \textcolor{gray}{0.9917} & \textcolor{gray}{0.9678} \\ \midrule
G4W+P4A & 4/8 & 0.3540 & 0.8225 & 0.5730 & 0.5018 & 0.5639 & 0.9849 & 0.9895 & 0.9651 \\
\textbf{Ours} & 4/8 & \textbf{0.3904} & \textbf{0.8475} & \textbf{0.5812} & \textbf{0.5160} & \textbf{0.6167} & \textbf{0.9859} & \textbf{0.9915} & \textbf{0.9687} \\
\bottomrule
\end{tabular}}
\vspace{-0.3cm}
\end{table*}


\section{Experiments}
\label{exp}
\subsection{Experimental Setup}

\noindent \textbf{Image generation.} We first evaluate our Q-DiT on the image generation task, closely following the evaluation setting used in DiT~\citep{peebles2023scalable}. 
We use the pre-trained DiT-XL/2 models with image resolutions of 256$\times$256 and 512$\times$512, converting them to FP16 as our full precison baseline model. 
For fast and accurate sampling, we adopt the DDIM sampler~\citep{songddim} with 50 and 100 sampling steps. Performance is also evaluated both with and without classifier-free guidance~\citep{ho2021classifier}. 
Note that "cfg" denotes the classifier-free guidance scale. We sample 10K images for both ImageNet 256$\times$256 and ImageNet 512$\times$512 in each setting,
and employ four metrics in our experiments, including Fréchet Inception Distance (FID)~\citep{heusel2017gans}, spatial FID (sFID)~\citep{salimans2016improved}, Inception Score (IS), and Precision. 

\noindent \textbf{Video generation.} We also evaluate Q-DiT on video generation, with the STDiT3 model from the Open-Sora project \citep{opensora}. Specifically, we sample five 2-second videos for each prompt in the VBench prompt suite~\citep{huang2023vbench} using a 30-step rectified flow scheduler with a cfg scale of 7.0. The results are assessed across 16 dimensions provided by VBench.

\noindent \textbf{Baselines.} We compare Q-DiT with five strong baselines:
\begin{enumerate}[label=\arabic*)]
    \item PTQ4DM~\citep{shang2023ptqdm}: A method specifically designed for UNet-based diffusion models, focusing on calibration of activations.
    \item RepQ-ViT~\citep{li2023repq}: A technique developed for the quantization of ViTs, aiming to reduce quantization errors in transformer activations.
    \item TFMQ-DM~\citep{Huang_2024_CVPR}: A PTQ framework specifically developed for diffusion models to preserve temporal features during quantization.
    \item PTQ4DiT~\citep{wu2024ptq4dit}: A tailored PTQ approach for DiTs that addresses the quantization challenges through CSB and SSC.
    \item G4W+P4A: A robust baseline we build in this work for both video and image generation tasks, utilizing GPTQ~\citep{frantar2022gptq} for weight quantization and PTQ4DM for activation quantization.

\end{enumerate}

\noindent \textbf{Others.} Q-DiT applies asymmetric quantization for both weights and activations, and uses GPTQ~\citep{frantar2022gptq} for weight quantization. A default group size of 128 is adopted, with optimal group size allocation for each layer determined through evolutionary search. The search space for group size $\mathcal{S}_g$ is \{32, 64, 128, 192, 288\}. Note that the group size for weights and activations within the same layer should be the same.

\begin{figure*}[t!]
    \centering
    \includegraphics[width=\textwidth]{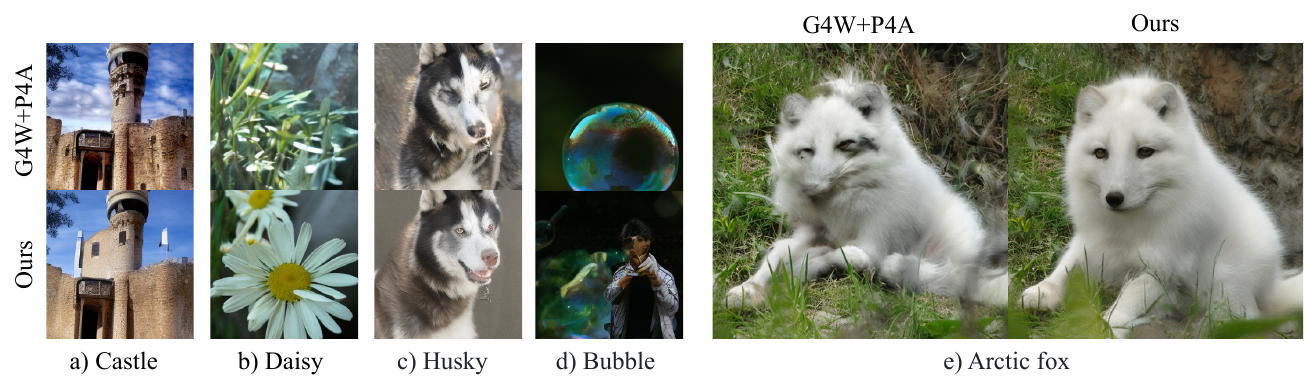}
  \caption{Qualititive results. Samples generated by G4W+P4A (one of our baselines) and Q-DiT with W4A8 on ImageNet 256$\times$256 and ImageNet 512$\times$512. For each example (a-e), the image generated by G4W+P4A shows notable artifacts and distortions. In contrast, our method produces cleaner and more realistic images, with better preservation of textures.
  }
  \label{fig: comparison}
  \vspace{-0.3cm}
\end{figure*}

\subsection{Main Results} 
\noindent \textbf{Image generation results.} The quantitative results for image generation are shown in Tab.~\ref{tab: 256_performance}. Specifically, in experiments conducted on ImageNet at a resolution of 256×256, the PTQ4DM, RepQ-ViT, TFMQ-DM, PTQ4DiT and G4W+P4A methods exhibit significant performance degradation at a bit-width of W6A8. In contrast, Q-DiT shows marked improvements over PTQ4DM under the W6A8 setting, effectively minimizing the impact of quantization. Notably, Q-DiT is closely aligned with the full-precision configuration, achieving an FID of 12.21 and an IS of 117.75. These results highlight the effectiveness of our method in achieving near-lossless compression in the W6A8 quantization setting.
When the bit-width is reduced to W4A8, the performance disparities among the methods become more pronounced. In particular, the other five baselines have severe performance degradation, while our method substantially outperforms them, dramatically reducing quantization loss with an FID of 15.76 and an IS of 98.78. This demonstrates a significant preservation of quality and diversity at lower bit-widths, highlighting the robustness of our approach under stringent quantization constraints.
Across varying steps (100 and 50) and classifier-free guidance scales, our method consistently shows superior performance, closely emulating the full-precision model metrics. The evaluation on the ImageNet 512$\times$512 dataset demonstrates consistent trends with the 256$\times$256 dataset, indicating that Q-DiT can also perform well in high-resolution image generation. Visual demonstrations in Fig.~\ref{fig: comparison} further illustrate that our method maintains superior image generation quality compared to the baseline.

\noindent \textbf{Video generation results.} Tab.~\ref{tab:video_model_performance} shows the results for video generation. Under a stringent W4A8 quantization setting, our method consistently outperforms G4W+P4A in 15 out of 16 metrics, exhibiting minimal degradation compared to the full-precision model. This indicates that our method performs well in terms of preserving video quality and maintaining video-condition consistency.

\subsection{Ablation Studies} 

To evaluate the effectiveness of each proposed component, we conduct ablation studies on ImageNet 256×256 with the DiT-XL/2 model.
The sampling steps and classifier-free guidance scale are set to 100 and 1.5, as detailed in Tab.~\ref{tab:ablation}.

\noindent \textbf{Incremental analysis of Q-DiT.} 
We begin our assessment with a round-to-nearest (RTN) baseline, which simply rounds weights and activations to the nearest available quantization level. Under the W4A8 configuration, RTN demonstrates significantly low performance across all metrics. Enhancing RTN by adjusting the quantization granularity to a group size of 128 markedly improves the results. The introduction of dynamic activation quantization led to a significant boost in generation quality, evidenced by an FID of 6.64, an sFID of 19.29, and an IS of 211.27. By further incorporating group size allocation, our approach achieves an impressive FID of 6.40, approaching the performance of the full-precision model.

\noindent 
\textbf{Comparisons of dynamic activation quantization methods.}
We also conducted experiments on activation quantization, as shown in Tab.~\ref{tab:W16A8}. Both TFMQ-DM and our method are quantized to W16A8 to isolate and compare the impact of activation quantization on overall performance. Our method achieves an FID of 5.34, demonstrating a significant improvement over TFMQ-DM, which has an FID of 7.74.
This highlights the effectiveness of our sample-aware dynamic activation quantization in maintaining model accuracy while reducing performance degradation compared to TFMQ-DM.

\noindent
\textbf{Comparisons of search methods.}
Furthermore, we also evaluate the effectiveness of the proposed search method (Alg. 1) used in group quantization and show the results in Tab.~\ref{tab:search}. We can find the proposed method significantly performs better than ILP method~\citep{moon2024instance}, Hessian-based search method~\citep{li2021brecq}, and the baseline, which demonstrates the effectiveness the our method.

\begin{table}[t]
\caption{Incremental analysis of individual components in our proposed method under the W4A8 setting.} 
\label{tab:ablation}
\resizebox{\columnwidth}{!}{
{%
\small
\begin{tabular}{@{}llllc@{}}
\toprule
~                     & FID $\downarrow$     & sFID $\downarrow$   & IS $\uparrow$ &      \\ \midrule
FP (W16A16)
&\textcolor{gray}{5.31}           &\textcolor{gray}{17.61}         &\textcolor{gray}{245.85}                 \\ \midrule
W4A8 RTN                                       &225.50             &88.54        &2.96                \\
+ Group size 128                          &13.77          &27.41         &146.93                 \\

\makecell[l]{\quad + Sample-wise Dynamic \\ \hspace{1em} \quad activation quantization}              & 6.64
& 19.29   & 211.27   \\ 

\makecell[l]{\quad \quad + Automatic quantization \\ \hspace{1em}\quad \quad granularity allocation}              & \textbf{6.40} & \textbf{18.60} & \textbf{211.72}  \\ \bottomrule
\end{tabular}}}
\vspace{-0.3cm}
\end{table}

\begin{table}[t]
\caption{Comparisons of dynamic activation quantization methods with W16A8 setting. TFMQ-DM is a method for timestep-aware activation quantization, whereas our approach is both timestep-wise and sample-wise.}
\centering
\label{tab:W16A8}
{%
\small
\begin{tabular}{ccccc}
\toprule
Method                     & FID $\downarrow$     & sFID $\downarrow$   & IS $\uparrow$ & \begin{tabular}[c]{@{}c@{}}Precision$\uparrow$ \end{tabular}     \\ \midrule
FP (W16A16)                   &\textcolor{gray}{5.31}           &\textcolor{gray}{17.61}         &\textcolor{gray}{245.85}         &\textcolor{gray}{0.8077}         \\ \midrule
TFMQ-DM            &7.74          &19.23         &204.56        &0.7765          \\
\textbf{Ours}      & \textbf{5.34} & \textbf{17.44} & \textbf{245.24} & \textbf{0.8048} \\ \bottomrule
\end{tabular}}
\vspace{-0.3cm}
\end{table}

\begin{table}[t]
\caption{ Comparisons of the proposed search method and potential counterparts.}
\centering
\label{tab:search}
{%
\small
\begin{tabular}{ccccc}
\toprule
Search method                     & FID $\downarrow$     & sFID $\downarrow$   & IS $\uparrow$ & \begin{tabular}[c]{@{}c@{}}Precision$\uparrow$ \end{tabular}     \\ \midrule
Group size = 128                  &\textcolor{gray}{6.64}           &\textcolor{gray}{19.29}         &\textcolor{gray}{211.27}         &\textcolor{gray}{0.7548}         \\ \midrule
ILP            &6.71          &19.20         &205.54        &0.7538          \\
Hessian-based                           &7.38          &19.41         &197.48        &0.7385          \\
\textbf{Ours}                          & \textbf{6.40} & \textbf{18.60} & \textbf{211.72} & \textbf{0.7609} \\ \bottomrule
\end{tabular}}
\vspace{-0.3cm}
\end{table}

\section{Conclusion}
Our study presents Q-DiT, a novel post-training quantization framework designed for DiTs.
To address the significant spatial variance of weights and activations in input channels, we introduced an automatic quantization granularity allocation method. Furthermore, to manage variations in activation ranges across different timesteps, we implemented dynamic activation quantization that adaptively adjusts quantization parameters during runtime. 
Extensive experiments have underscored the effectiveness of our approach, showcasing its superiority over existing baselines. Notably, even when quantizing the model to W4A8 on the ImageNet $256\times256$ dataset, the FID increased by only 1.09.

\noindent \textbf{Limitations and future work.}
One of the primary limitations of the current Q-DiT approach is its reliance on evolutionary algorithms to determine the optimal group size configuration for quantization. This process is computationally expensive and time-consuming, increasing the overall cost and duration of optimization. We plan to optimize this part in the future work.

{
    \small
    \bibliographystyle{ieeenat_fullname}
    \bibliography{main}
}

\end{document}